\documentclass[conference]{IEEEtran}
\IEEEoverridecommandlockouts

\usepackage{cite}
\usepackage{amsmath,amssymb,amsfonts}
\usepackage{algorithmic}
\usepackage{graphicx}
\usepackage{textcomp}
\usepackage{xcolor}
\usepackage{comment}

\def\BibTeX{{\rm B\kern-.05em{\sc i\kern-.025em b}\kern-.08em
    T\kern-.1667em\lower.7ex\hbox{E}\kern-.125emX}}
\begin{document}

\title{Knowledge Distillation in Deep Learning and its Applications\\}

\author{\IEEEauthorblockN{Abdolmaged Alkhulaifi\IEEEauthorrefmark{1} \textsuperscript{\textsection},
Fahad Alsahli\IEEEauthorrefmark{2} \textsuperscript{\textsection},
and Irfan Ahmad\IEEEauthorrefmark{3}}
\IEEEauthorblockA{Department of Information and Computer Science\\
King Fahd University of Petroleum and Minerals\\
Dhahran, Saudi Arabia\\
\IEEEauthorrefmark{1}g201905030@kfupm.edu.sa,
\IEEEauthorrefmark{2}g201221720@kfupm.edu.sa,
\IEEEauthorrefmark{3}irfan.ahmad@kfupm.edu.sa}}

\maketitle

\begingroup\renewcommand\thefootnote{\textsection}
\footnotetext{Equal contribution}
\endgroup

\begin{abstract}

Deep learning based models are relatively large, and it is hard to deploy such models on resource-limited devices such as mobile phones and embedded devices. One possible solution is knowledge distillation whereby a smaller model (student model) is trained by utilizing the information from a larger model (teacher model). In this paper, we present a survey of knowledge distillation techniques applied to deep learning models. To compare the performances of different techniques, we propose a new metric called distillation metric. Distillation metric compares different knowledge distillation algorithms based on sizes and accuracy scores. Based on the survey, some interesting conclusions are drawn and presented in this paper.\\
\end{abstract}


\section{Introduction}
Deep learning has succeeded in several fields such as Computer Vision (CV) and Natural Language Processing (NLP). This is due to the fact that deep learning models are relatively large and could capture complex patterns and features in data. But, at the same time, large model sizes lead to difficulties in deploying them on end devices. \\

To solve this issue, researchers and practitioners have applied knowledge distillation on deep learning approaches. It should be emphasized that knowledge distillation is different from transfer learning. The goal of knowledge distillation is to provide smaller models that solve the same task as the larger models \cite{KD-paper}, whereas the goal of transfer learning is to reduce training time of models that solve a task similar to the task solved by some other model (cf. \cite{transfer-learning}). Knowledge distillation accomplishes its goal by altering loss functions of models being trained (student models) to account for output of hidden layers of pre-trained models (teacher models). On the other hand, transfer learning achieves its goal by initializing parameters of a model by learnt parameters of a pre-trained model. \\

There are many techniques presented in the literature for knowledge distillation. As a result, there is a need to summarize them so that researchers and practitioners could have a clear understanding of the techniques. Cheng et al. \cite{survey} provided a survey in 2017. The survey was not specific to knowledge distillation, and it considered several model compression approaches such as parameter pruning and transferred/compact convolutional filters. So, the survey was limited in terms of the coverage of knowledge distillation approaches 

The limited coverage motivated us to present a comprehensive survey on recent knowledge distillation methods.\\

Since there are many proposed knowledge distillation methods, they should be compared appropriately. Knowledge distillation approaches can be compared by several metrics such as reductions in model sizes, accuracy scores, processing times, and so on. Our main criteria are reduction in model sizes and the change in accuracy scores. Accordingly, we propose a metric--termed distillation metric--that takes into account the two criteria. 
Distillation metric is presented in section \ref{distillation-metric}. \\


The rest of the paper is organized as follows: In Section II, we provide a background on knowledge distillation. In section III, we present and discuss our proposed distillation metric. Section IV contains the surveyed approaches. We provide our discussion on surveyed approaches and an outlook on knowledge distillation in section V. Finally, we present our conclusions in section VI.

\section{Background}
The main goal of knowledge distillation is to produce smaller models (student models) to solve the same task as larger models (teacher models) with the condition that developed models should perform better than baseline models of the same size and trained without teacher models \cite{KD-paper}. Knowledge distillation aims to train student models by having them to mimic behaviours of teacher models. The process is achieved by comparing student's activations (e.g., outputs of hidden layers) to teacher's. There are no conventions that guide student models' sizes. For example, two practitioners might have student models with different sizes although they use the same teacher model. This situation is caused by different requirements in different domains. For example, maximum allowed model size on some device. \\ 

There exist some knowledge distillation methods that target teacher and student networks having the same size(e.g., \cite{f1}). In such a case, the purpose of knowledge distillation would be to accelerate training time. Although an algorithm is developed to distill knowledge from a teacher model to a student model having the same sizes, the same algorithm might be used to distill knowledge from a teacher to a smaller student. This is because, based on our survey, there is no restriction on model sizes, and it is up to model designers to map teacher's activations to student's. So, in general settings, knowledge distillation is utilized to provide smaller student models that have good maintainability of their teacher models' accuracy scores. \\

Consequently, one could compare different knowledge distillation algorithms by their reductions in model sizes. In addition, algorithms might be compared by how much accuracy they maintain as compared to teacher models. There is no rule that governs how much reduction is best for all cases. For instance, if one needs to apply a knowledge distillation algorithm, they need to compare the algorithm's performance, in terms of reductions in size and accuracy, to their system's requirements. Based on the requirements, they can decide which algorithm best fits their situation. To ease the process of comparison, we develop distillation metric which compares knowledge distillation algorithms based on model sizes and accuracy scores. For a detailed description, please refer to section \ref{distillation-metric}. \\

There are different knowledge distillation approaches applied to deep learning models. For example, there exist approaches that distill knowledge from a single teacher to a single student. Also, other approaches distill knowledge from several teachers to a single student. Knowledge distillation could also be applied to provide an ensemble of student networks. In section \ref{survey-sec}, we present recent knowledge distillation approaches that are applied on deep learning based architectures. \\

\section{Distillation Metric} \label{distillation-metric}
We propose distillation metric to compare different knowledge distillation methods. The metric considers ratios of student network's size (first ratio) and accuracy score (second ratio) to teacher's. To have a good reduction in size, first ratio should be as small as possible. For a distillation method to have a good maintainability of accuracy, second ratio should be as close to 1 as possible. To satisfy these requirements, we develop the following equation:

\begin{equation} \label{dm-eq}
DS = \alpha * (\frac{student_s}{teacher_s}) + (1-\alpha) * (1 - \frac{student_a}{teacher_a})
\end{equation}

where $DS$ stands for distillation score, $student_s$ and $student_a$ are student size and accuracy respectively, and $teacher_s$ and $teacher_a$ are teacher size and accuracy respectively. Parameter $\alpha \in [0, 1]$ is a weight to indicate importance of first and second ratio, i.e., size and accuracy. The weight is assigned by distillation designers based on their system's requirements. For example, if some system's requirements prefer small model sizes over maintaining accuracy, designers might have $\alpha > 0.5$ that best satisfies their requirements. In this paper, we have equal weights of 0.5 for both ratios as size and accuracy reductions are equally important for the scope of this survey. \\

It should be noted that when student accuracy is better than teacher's, then second ratio would be greater than 1. This causes the right operand of the addition operation (i.e., 1 - second ratio) to evaluate to a negative value. Hence, DS is decreased, and it could be less than zero especially if weight of second ratio is larger. This is a valid result since it indicates a very small value of first ratio compared to second ratio. On other words, this behaviour indicates a large reduction in model size while providing better accuracy scores than teacher model at the same time. As presented in section \ref{survey-sec}, a student model with a better accuracy is not a common case. It could be achieved, for example, by having an ensemble of student models. \\

Regarding behaviour of distillation metric, it is as the following: The closer distillation score to 0 is, the better the knowledge distillation is. To illustrate, an optimal knowledge distillation algorithm would provide a value that is very close to 0 for first ratio (e.g., student size is very small compared to teacher's), and it would produce a value of 1 for second ratio (e.g., student and teacher networks have the same accuracy score). As a result, distillation score approaches 0 as the first ratio approaches 0, and the second ratio approaches 1.

\section{Survey} \label{survey-sec}
This section includes recent works that target knowledge distillation in deep learning. It is divided into two sub-sections. First section considers work that utilizes soft labels of teachers to train students. Soft labels refers to the output of hidden layers. Second section considers work that performs some transformation on soft labels of teachers before utilizing them to train students. In this survey, our main criteria are reductions of sizes and accuracy scores of student models against the corresponding teacher models. Some researchers did not provide these details. As a result, our survey is missing them. Regarding distillation scores for surveyed work, they are presented in tables \ref{summaryTable} and \ref{summaryTable2}. This is because some researchers evaluated their approaches on different datasets, and some other researchers had several variants of their methods. We only consider best results to compute distillation scores. Hence, we mention distillation score for each paper on the tables to avoid confusion.

\subsection{Techniques Using Soft Labels to Train Student Models}
Every neural network can be divided into two parts: feature extractor (backbone) and data adapter (task-head). Backbone network encodes input data into a feature map that represents input data while the task-head uses the feature map to perform a specific task, \textit{e.g.,} detecting a certain object in an image. While other approaches try to train a student model to mimic a teacher model, Gao et al. \cite{a5} proposed to only train the backbone of a student model to mimic the feature extraction output of a teacher model. After that, the student model is trained on ground truth data while fixing parameters on the backbone layers. The knowledge distillation process only happened during training of the backbone layers of the smaller student model, which allowed it to be trained on different dataset than the teacher model. Gao et al. conducted experiments on CIFAR-100 \cite{CIFAR-10} and ImageNet \cite{imagenet} datasets for image classification and CASIA-WebFace \cite{CASIA} and IJB-A \cite{IJB-A} datasets for facial recognition. Using ResNet-34 \cite{ResNet} model structure as a teacher (with 21.282 million parameters) and ResNet-18 as a student (with 11.174 million parameters), they achieved an accuracy of 70.77\% for student model, 71.21\% for teacher model, and 67.96\% accuracy for baseline model (trained without distilling knowledge from teacher model) on CIFAR-100 dataset. While for ImageNet, student model achieved 71.36\%, teacher got 73.55\%, and baseline got 69.57\% . On the other hand, they used the facial recognition task to evaluate the generalization ability of their approach by using a teacher trained on CASIA-WebFace dataset to train the backbone of a student and then fitting the student model on IJB-A dataset. Student model achieved 80.5\% accuracy in the verification task compared to teacher 83.7\% and 76.9\% accuracy of baseline model trained directly on IJB-A dataset. These result showed the capability of their approach in allowing student models to generalize and fit over other datasets by only learning the backbone of teacher models trained on different dataset.\\

While deep learning has achieved great success across a wide range of domains, it remains difficult to identify the reasoning behind model predictions, especially if models are complex. To tackle this issue, Liu et al. \cite{a6} proposed a method of converting deep neural networks to decision trees via knowledge distillation. The proposed approach consisted of training a Convolutional Neural Network (CNN) first with the given dataset. Using the feature set from the training dataset as input and the logits from the trained model as output, they trained a classification and regression trees (CART) model, where logits are scores before the SoftMax activations. They applied their approach on MNIST \cite{mnist} and connect-4 \cite{connect-4} datasets where a deep CNN was used for MNIST, and multilayer perceptron was used for connect-4. For both tasks, the student decision tree model that was trained  by a deep learning teacher model achieved between 1-5\% better accuracy than a decision tree model with similar depth trained without a deep learning teacher model. However, accuracy scores of student and teacher models on MNIST were 86.55\% and 99.25\% respectively. While for connect-4, student model achieved 73.42\%, and teacher model achieved 86.62\%.\\

In \cite{f9}, Furlanello et al. proposed an ensemble knowledge distillation method called Born-Again Neural Networks. The method considered the issue of teacher and student models having the same architecture. The method first trained a teacher model normally. Then, it trained a student model using training labels and teacher’s predictions. After that, it trained a second student model using training labels and previous student’s predictions, and so on. For instance, $student_i$ was trained by utilizing training labels and predictions of $student_{i-1}$ for $i \in [1, n]$, where $n$ is the number of student models. When student models were used for prediction, their results were averaged. Furlanello et al. claimed that the method would produce better models since it was based on ensemble models, and a model was trained on training labels and predictions of a previously trained model. To validate their claim, Furlanello et al. trained a DenseNet \cite{DenseNet} as a teacher model. They applied their method using 3 students. As training data, they utilized CIFAR-100. Teacher model got an error score of 16.87\% (83.13\% accuracy score), and student models got an error score of 14.9\% (85.1\% accuracy score). So, the distillation method resulted in a 11.67\% decrease in error score (2.369\% increase in accuracy score). \\

Polino et al. \cite{f10} developed a knowledge distillation approach for quantized models. Quantized models are models whose weights are represented by a limited number of bits such as 2-bit or 4-bit integers. Quantized models are used to develop hardware implementations of deep learning architectures as they provide lower power consumption and lower processing times compared to normal models (full-precision models) \cite{quant1}. The distillation approach had 2 variants. First variant was called quantized distillation, and it trained a quantized student model and a full-precision student model. The two models were trained according to true labels and teacher’s predictions. The main purpose of full-precision model was to compute gradients and update quantized model accordingly. As claimed by Polino et al., the reason behind this process was that there was no objective function that accounted for quantized weights. This issue motivated Polino et al. to develop the second variant of their knowledge distillation approach, and they called it differentiable quantization. They defined an objective function to address the issue of quantized weights. As a result, there would be no need for full-precision student model. To test their approach, Polino et al. defined a teacher model with 36.5M parameters and a student model with 17.2M parameters. The models were wide residual networks \cite{Wide-residual-networks}, and they were trained using CIFAR-100 data. Teacher model got an accuracy score of 77.21\%. Student trained using quantized distillation got an accuracy score of 76.31\%. Student trained using differentiable quantization got an accuracy score of 77.07\%. So, quantized distillation resulted in a 1.165\% decrease in accuracy score, whereas differentiable quantization resulted in a 0.1813\% decrease in accuracy score. Both variants provided a 52.87\% decrease in model size. \\

Wang et al. \cite{f6} proposed a distillation method that trained a student network by comparing its soft labels to a teacher’s labels. Teacher and student networks were U-Net \cite{U-Net} models. Teacher had 2.87M parameters, while student had 603K parameters. Janelia \cite{Janelia} data was utilized for training. The distillation achieved a reduction of 78.99\% in model size. \\

Kurata and Audhkhasi \cite{f8} developed a distillation approach that targeted sequence models for speech recognition. The distillation goal was to transfer knowledge of a Bidirectional Long Short-Term Memory (BiLSTM) model to an LSTM model. This was achieved by considering teacher's soft labels and comparing outputs of three time steps of teacher network to a single time step output of student network. Teacher network had 138M parameters, and student network had 62M parameters. For training, two subsets of NIST Hub5 2000 evaluation data \footnote{https://catalog.ldc.upenn.edu/LDC2002T43} were utilized. They were Switchboard (SWB) with 21.4K words and CallHome (CH) with 21.6K words. The authors utilized Word Error Rate (WER) score as a performance metric. Teacher network could get WER scores of 13.4\% (86.6\% accuracy) on SWB data and 23.2\% (76.8\% accuracy) on CH data. On the other hand, student network could get WER scores of 15.7\% (84.3\% accuracy) on SWB data and 27.8\% (72.2\% accuracy) on CH data. WER increase on SWB data was 17.16\% (2.655\% accuracy reduction), whereas it was 19.83\% (5.989\% accuracy reduction) on CH data. The distillation approach could result in a reduction of 55.07\% in model size. \\

Mun’im et al. \cite{f7} developed a distillation approach that targeted Seq2Seq \cite{Seq2Seq} models for speech recognition. The approach trained a student network via teacher’s soft labels and k-best hypotheses (e.g., k-best outputs) where k was a hyper-parameter, and it was independent from model sizes. The authors developed a teacher network with 16.8M parameters, a student network with 6.1M parameters (student-mid), and a student network with 1.7M parameters (student-small). The networks were trained using Wall Street Journal (WSJ) corpus \footnote{https://catalog.ldc.upenn.edu/LDC93S6B}, and their accuracies were measured using word-error rate (WER) scores. Teacher got a WER score of 15.3\% (84.7\% accuracy). Student-mid (with k=10) got a WER score of 19.7\% (80.3\% accuracy), so the increase in WER was 28.76\% (5.195\% accuracy reduction). Student-small (with k=5) got a WER score of 22.3\% (77.7\% accuracy), so the increase in WER was 45.75\% (8.264\% accuracy reduction). The reduction in size for student-mid was 63.69\%, while it was 89.88\% for student-small. In addition, Mun’im et al. compared student-mid and student-small models against 2 baseline models. Baseline models had the same sizes as their corresponding student models, and they were trained without teacher models. Mun’im et al. did not provide k values for baseline models. Student-mid's baseline got a 21.8\% WER score (78.2\% accuracy), and student-small's baseline got a 28.7\% WER score (71.3\% accuracy). Student-mid had a 9.63\% WER reduction (2.68\% accuracy increase) compared to its baseline, whereas student-small had a 22.2\% WER reduction (8.97\% accuracy increase) compared to its baseline. \\

Training a compact student network to mimic a well-trained and converged teacher model can be challenging. The same rationality can be found in school-curriculum, where students at early stages are taught easy courses and further increasing the difficulty as they approach later stages. From this observation, Jin et al. \cite{a8} proposed that instead of training student models to mimic converged teacher models, student models were trained on different checkpoints of teacher models until teacher models converged. For selecting checkpoints, a greedy search strategy was proposed that finds efficient checkpoints that are easy for the student to learn.  Once checkpoints were selected, a student model's parameters were optimized sequentially across checkpoints, while splitting data used for training across the different stages depending on it’s hardness defined by a hardness metric that was proposed by the authors. Experiments were conducted using ResNet-50 \cite{ResNet} network as a teacher (with 23.521 million parameter) and a compact MobileNetV2 \cite{mobilenet} network as a student (with 3.4 million parameter). Student model achieved 70.85\% accuracy on CIFAR-100 dataset compared to teacher 79.34\% and baseline 61.88\%. While on ImageNet \cite{imagenet} dataset, student model achieved 68.21\% compared to teacher 75.49\% and 64.2\% achieved by baseline model.\\

Unlike other knowledge distillation methods where neuron responses of teacher model is the focus when transferring knowledge to students, Heo et al. \cite{a9} proposed to focus on transferring activation boundaries of teacher instead. Activation boundary is a hyperplane that decides whether the neurons are active or not. In Pan et al. \cite{a9.1}, decision boundary of neural network classifier was proven to be a combination of activation boundaries, which made them an important knowledge to be transferred to student model. Based on this, Heo et al. proposed an activation transfer loss that penalized when neurons activations of teacher and student were different in hidden layers. Since both teacher and student model, most likely, would not have the same number of neurons, Heo et al. utilized a connector function that converts the vector of neurons of student model to the same size of the vector of neurons in teacher model. By applying the proposed loss function, activation boundaries of teacher model were transferred to student model. To evaluate their approach, they conducted an experiment using MIT scene classifier \cite{mit_scenes} dataset. For teacher model, ResNet50 \cite{ResNet} model was used (with 23.521 million parameters) and it was trained on ImageNet \cite{imagenet} dataset. For student, small MobileNet \cite{mobilenet} (with 3.4 million parameters) model was used. In MIT dataset, their student model achieved 74.10\% accuracy compared to teacher 69.78\% and baseline 64.93\%. This showed that their method improved the performance of a model with transferred knowledge from a pretrained model on different dataset while fitting on a new dataset.\\

Table \ref{summaryTable} provides a summary of presented work. It shows that best approach in terms of size reduction is proposed by Mun’im et al. \cite{f7} with a reduction of 89.88\% in size. Mun’im et al. also got the best distillation score. However, the table shows that best approach in terms of maintaining accuracy is proposed by Heo et al. \cite{a9} with an increase in accuracy of 6.19\%.

\begin{table*}[ht]
\centering
\caption{\label{summaryTable}Summary of knowledge distillation approaches that utilize soft labels of teachers to train student model. In case of several students, results of student with largest size reduction are reported. In case of several datasets, dataset associated with lowest accuracy reduction is recorded. Baseline models had the same sizes as student models, but they were trained without teacher models.}

\begin{tabular}{|p{0.14\linewidth}|l|p{0.14\linewidth}|p{0.15\linewidth}|p{0.15\linewidth}|p{0.07\linewidth}|p{0.07\linewidth}|}
\hline
Reference               & Targeted Architecture & Utilized Data     & Reduction in Accuracy Compared to Teacher & Improvement in Accuracy Compared to Baseline & Reduction in Size & Distillation Score \\ \hline

Gao et al. \cite{a5}         & CNN                   & CIFAR-100  & 0.618\%  & 4.135\%                  & 50\%    &    0.253   \\ \hline

Liu et al. \cite{a6}         & Decision tree                   & MNIST & 12.796\%     & 1-5\%               & -   &    -  \\ \hline

Furlanello et al. \cite{f9}          & DenseNet                & CIFAR-100    & 2.369\% (increase) & -      & -    &     -  \\ \hline

Polino et al. \cite{f10}          & Wide ResNet   & CIFAR-100    &   0.1813\%  &   -     & 52.87\%    &    0.2365   \\ \hline

Wang et al. \cite{f6}          & U-Net                 & Janelia    & -   & -        & 78.99\%     &   -   \\ \hline

Kurata and Audhkhasi \cite{f8} & LSTM                  & SWB               & 2.655\%   & -             & 55.07\%    &    0.2379   \\ \hline

Mun’im et al. \cite{f7}        & Seq2Seq               & WSJ               & 8.264\% & 8.97\%  & 89.88\%      &   0.09192  \\ \hline

Jin et al. \cite{a8}          & CNN                   & ImageNet   & 9.644\%      & 6.246\%               & 70.66\%      &   0.1949  \\ \hline

Heo et al. \cite{a9}         & CNN                   & ImageNet to MIT scene,  & 6.191\% (increase)  & 14.123\%         & 70.66\%      &   0.1157  \\ \hline

\end{tabular}
\end{table*}

\subsection{Techniques Using Transformation of Soft Labels to Train Student Models}
Lopes et al. \cite{a1} proposed that instead of using the original dataset used to train a teacher for transferring knowledge to a student model, a metadata which holds a summary of activations of the teacher model during training on the original dataset. The metadata includes top layer activation statistics, all layer’s activation statistics, all-layers spectral activation record, and layer-pairs spectral activation record. Then using one of the collected metadata, we can capture the view of the teacher model of the dataset and hence we can reconstruct a new dataset that can be used to train a compact student model. Experiments were conducted using a fully connected network and convolution network on MNIST and convolution network on CelebA \cite{celeba} dataset. The authors trained a compact student model that has 50\% less parameter than the teacher model for both tasks. The results of their experiments concluded that their distillation approach using layer-pairs spectral activation record to reconstruct the dataset achieved better results than the other activation records. For MNIST, they achieved an accuracy of 91.24\% compared to the 96.95\% achieved by the teacher model with the fully connected network and 92.47\% accuracy by the student model compared to the 98.91\% accuracy by the teacher model with CNN models. While for CelebA, their student model achieved an accuracy of 76.94\% compared to the 80.82\% achieved with the teacher model with convolution neural network. \\

When tackling problems where only few samples are available, it can make models overfit easily. Kimura et al. \cite{a2} proposed a method that allowed training networks with few samples while avoiding overfitting using knowledge distillation. In their approach, they first trained a reference model with few samples using Gaussian processes (GP) instead of neural network. Then, the samples used for training were augmented using inducing point method via iterative optimization. Finally, the student model was trained with the augmented data using loss function defined in the paper with the GP teacher model to be imitated by the student model. The authors ran their experiment against MNIST and fashion-MNIST \cite{fashionMNIST} with a 3-layer CNN as a student model. By using only 1.25K samples and augmented to 10K, their proposed method achieved an accuracy of 44.1\% and 44.8\% for MNIST and fashion MNIST respectively compared to 37.9\% and 39.3\% achieved via training on them directly with the same model. While the Gaussian processes teacher model achieved 39.9\% and 44.6\% on MNIST and fashion MNIST respectively. \\

In order to transfer knowledge from a teacher model and to train a student model, we need access to the dataset that was used to train the teacher model, or some metadata as was described by Lopes et al. \cite{a1}. Nayak et al. \cite{a3} proposed a method to train the student model without using any dataset or metadata. The method worked by extracting data from the teacher model through modeling the data distribution in the SoftMax space. Hence, new samples could be synthesized from the extracted information and used to train the student model. Unlike generative adversarial networks (GANs) where they generates data that is similar to the real data (by fooling a discriminative network), here the synthesized data were generated based on triggering the activation of the neurons before the SoftMax function. They applied their framework on CNN models and used MNIST , Fashion MNIST and CIFAR-10 datasets to compare their approach with other knowledge distillation approaches. The student models that were used have about 40\% less parameters compared to the teacher model. The trained student model achieved accuracy of 98.77\% compared to the teacher models 99.34\% in MNIST. While in fashion MNIST and CIFAR-10 archiving 79.62\% and 69.56\% respectively compared to the 90.84\% and 83.03\% achieved by the teacher model respectively. \\

Yim et al. \cite{f1} proposed a two-stage distillation for CNNs. The first stage was to define two matrices between activations of two layers, and the layers had not to be consecutive. The first matrix corresponded to teacher network, and the second matrix corresponded to student network. Then, the student was trained to mimic the teacher’s matrix. After that, the second stage began by training the student normally. Yim et al. considered CIFAR-10 data for training. Both teacher and student networks had the same number of layers of 26. Training teacher network took 64K iterations, whereas training student network took 42K iterations (e.g., 21K for the first stage and 21K for the second stage). Teacher’s accuracy was 92\%, and student's accuracy was 92.28\%. The distillation method resulted in a reduction of 34.37\% in training time and an increase in accuracy of 0.3043\%.  \\


Fukuda et al. \cite{a4} proposed a knowledge distillation approach by training a student model using multiple teacher models. The concept of using multiple teachers to train a single student model was already explored in previous studies. In previous approaches,  the output of the teacher models trained on different data or features are combined into one output distribution and then training the student model using that distribution \cite{KD-paper} \cite{a4.2} \cite{a4.3}. The approach proposed by Fukuda et al. was to opt out of combining the teacher models output distribution and to train the student on the individual output distribution. The authors argued that this would help the student model to observe the input data from different angles and would help the model to generalize better. For the experiment, the authors used the aurora task 4 \cite{aurora}, which is a speech recognition experimental framework, and compared two state of the art models VGG and an LSTM model against a CNN student model that was trained by VGG and LSTM models. In their experiment they showed that the compact model achieved a word error of 11.2\% (88.8\% accuracy) compared to the LSTM model 11.7\% (88.3\% accuracy) and VGG model 10.5\% (89.5\% accuracy) in the speech recognition task. They also ran another experiment where they trained a compact CNN model (consist of 2 convolution layers of 64 and 128 nodes and 2 fully connected layers with 768 neurons in each layer) directly on the dataset and compared it with the same model trained with two teachers (VGG and LSTM), the baseline model achived  a word error of 15.1\% (84.9\% accuracy) compared to the student model 13.2\% (86.8\% accuracy).\\

Pintea et al. \cite{f5} developed a knowledge distillation approach that mapped several teacher blocks to a single student block. The mapping was achieved via recurrence relations. Pintea et al. proposed three variants of the approach and trained a student network for each variant. So, there were three student networks. The authors considered CIFAR-10 dataset. Teacher network was ResNet \cite{ResNet} with 1.235M parameters, and it achieved an accuracy score of 93.28\%. Three student networks were ResNet models having 122K, 122K, and 73K parameters respectively. Their respective accuracy scores were 89.81\%, 89.25\%, and 88.33\%. So, the first distillation method achieved a reduction of 90.12\% in number of parameters and 3.72\% in accuracy score. The second distillation method achieved a reduction of 90.12\% in number of parameters and 4.32\% in accuracy score. The third distillation method achieved a reduction of 94.09\% in number of parameters and 5.307\% in accuracy score \\

While other teacher-student knowledge distillation approaches relies on a pre trained teacher model to transfer the knowledge to a student model. Zhou et al. \cite{a7} proposed to train the teacher (booster net) and the student (lightweight net) together. This was done by sharing the backbone layers of the two models during training and then using a function where it contained the loss of the booster network, the loss of the lightweight network, and the difference in the logits of both networks. To prevent the objective function from hindering the performance of the booster network, a gradient block scheme was developed to prevent the booster network specific parameter from updating during the backpropagation of the objective function which would allow the booster network to directly learn from the ground truth labels. To improve their approach further, they used the knowledge distillation loss function from Hinton et al. \cite{KD-paper} in their objective function. The authors ran experiments on CIFAR-10 using lightweight network consisting of 0.2M parameters and a booster network consisting of 0.6M parameters. The experiments resulted in error rate of 7.52 (92.48\% accuracy) for lightweight model compared to 6.58 (93.42\% accuracy) for booster model and 8.77 (91.23\% accuracy) achieved by a baseline lightweight model. They also ran an experiment using CIFAR-100 and SVHN \cite{svhn} datasets using different model architecture but with the same number of parameters for both the lightweight and booster models. The lightweight model achieved an error rate of 2.20 (97.8\% accuracy) on SVHN compared to 3.58 (96.42\% accuracy) achieved by the base model. While on CIFAR-100, the lightweight model achieved an error rate of 33.0 (67\% accuracy) and the baseline model achieved an error rate of 43.7 (56.3\% accuracy). \\

He et al. \cite{f2} proposed a two stage distillation. The first stage composed of compressing (e.g., dimensionality reduction) a teacher’s knowledge space to a student’s space due to the difference in architecture between the two networks. Compression was done using an auto-encoder network. For the second stage, its goal was to capture long-term dependencies. This was needed because student network could not capture such dependencies due to its relatively small size. He et al. considered ResNet-50 \cite{ResNet} architecture and PASCAL \cite{PASCAL} data. Teacher network had 26.82M parameters while student network had 2.11M parameters. The authors utilized pixel Intersection Over Union (mIOU) as an accuracy metric. Teacher network got mIOU of 76.21\%, and student network got mIOU of 72.5\%. The distillation resulted in a reduction of 92.13\% in model size and 4.868\% of mIOU. \\


Wu et al. \cite{f4} developed a multi-teacher distillation framework. Knowledge was transferred to student by taking a weighted average of teachers’ knowledge. The framework targeted CNN network for action recognition. Teacher network was CoViAR \cite{CoViAR} which was composed of three ResNet \cite{ResNet} models. The models were ResNet-18 with 11.2M parameters, ResNet-18 with 11.2M parameters, and ResNet-152 with 58.2M parameters. So, teacher network had 80.6M parameters in total. Student network was composed of a ResNet-18 model with 33.6M parameters. Wu et al. considered UCF-101 \cite{UCF-101} and HMDB51 \cite{HMDB51} data for training. For UCF-101 data, teacher achieved accuracy of 90.29\% while student achieved accuracy of 88.50\%. The reduction in accuracy was 1.982\%. For HMDB51 data, teacher achieved accuracy of 56.51\% while student achieved accuracy of 56.16\%. The reduction in accuracy was  0.6193\%. The framework resulted in a reduction of 58.31\% in number of parameters. \\

Previous knowledge distillation approaches only considered the instance features (the soft output of the layer) to be transferred from the teacher model to the student model. This made it hard for student models to learn the relationship between the instance feature and the sample with different and compact model architecture. Liu et al. \cite{a10} proposed representing the knowledge using an instance relation graph (IRG). For each layer in the model, an IRG was created where vertices represent the instance features and edges represent the instance relationship. Transformation function was defined to transform two IRG of adjacent layers into new IRG which contained the feature space knowledge of the two layers. Using IRG of the teacher layers and student layers, a loss function was defined to help train the student model using the knowledge encapsulated in the IRG of the teacher. The approach was tested on CIFAR10, CIFAR100-coarse and CIFAR100-fine \cite{CIFAR-10} image classification datasets with the ResNet20 \cite{ResNet} model as a teacher (with 1.06M parameters) and a shrunk down ResNet20 model as a student (with about 0.28M parameters). The student model achieved 90.69\%, 74.64\% and 62.25\% accuracy on CIFAR10, CIFAR100-coarse and CIFAR100-fine respectively compared to the teachers 91.45\%, 78.40\% and 68.42\% respectively and compared to the baseline model 88.36\%, 72.51\% and 59.88\% respectively.\\

Table \ref{summaryTable2} provides a summary of presented work. It shows that best approach in terms of size reduction is proposed by Pintea et al. \cite{f5} with a reduction of 94.09\% in size. Also, Pintea et al. got the best distillation score. The table shows that best approach in terms of maintaining accuracy is proposed by Kimura et al. \cite{a2} with an increase in accuracy of 10.526\%.

\begin{table*}[ht]
\centering
\caption{\label{summaryTable2}Summary of knowledge distillation approaches that performs some transformation on soft labels of teacher models to be used for training the student model. In case of several students, results of student with largest size reduction are reported. In case of several datasets, dataset associated with lowest accuracy reduction is recorded. Baseline models had the same sizes as student models, but they were trained without teacher models.}

\begin{tabular}{|p{0.14\linewidth}|l|p{0.14\linewidth}|p{0.15\linewidth}|p{0.15\linewidth}|p{0.07\linewidth}|p{0.07\linewidth}|}
\hline
Reference               & Targeted Architecture & Utilized Data     & Reduction in Accuracy Compared to Teacher & Improvement in Accuracy Compared to Baseline & Reduction in Size & Distillation Score \\ \hline

Lopes et al. \cite{a1}         & CNN                   & MNIST  & 4.8\%   & 5.699\% (decrease)                & 50\%    &    0.274   \\ \hline

Kimura et al. \cite{a2}         & CNN                   & MNIST & 10.526\% (increase)           &   16.359\%      & -   &   -  \\ \hline

Nayak et al. \cite{a3}         & CNN                   & MNIST  & 0.57\%  & -  & 40\%    &   0.3028    \\ \hline

Yim et al. \cite{f1}          & CNN                   & CIFAR-10  & 0.3043\% (increase)           & -        & -     &   -   \\ \hline

Fukuda et al. \cite{a4}        & CNN                   & Aurora  & 0.782\%  &   2.238\%              & -    &   -    \\ \hline

Pintea et al. \cite{f5}       & ResNet                & CIFAR-10          & 5.307\% &-              & 94.09\%       &  0.05609  \\ \hline

 Zhou et al. \cite{a7}        & CNN                   & CIFAR-10 & 1.006\%  &  1.37\%               & 66\%     &  0.1717  \\ \hline
 
He et al. \cite{f2}           & ResNet                & PASCAL      & 4.868\% (mIOU)   & -       & 92.13\%     &   0.06368  \\ \hline

Wu et al. \cite{f4}           & ResNet                & HMDB51      & 0.6193\%    & -          & 58.31\%    &    0.2115   \\ \hline

Liu et al. \cite{a10}         & CNN                   & CIFAR10 & 0.831\% &   2.637\%            & 73.59\%     &   0.1362   \\ \hline

\end{tabular}
\end{table*}

\section{Discussion and Outlook}

Distillation scores in tables \ref{summaryTable} and \ref{summaryTable2} have an average value of 0.1804 and a standard deviation of 0.07835. There are 6 distillation scores better than average. Most of the distillation algorithms corresponding to these scores have more than 70\% size reductions and less than 5\% accuracy reductions. There are 7 distillation scores that are worse than average. Majority of their algorithms have size reductions ranging from 50\% to 60\%. Low size reductions cause algorithms to have relatively high distillation scores although some of them have less than 1\% accuracy reductions. For example, algorithm developed by Wu et al. \cite{f4} has a very good accuracy maintainability, but its size reduction is 58.31\%. By inspecting tables, we can see that a distillation score of less than 0.2 would generally correspond to an algorithm having good size reduction and accuracy maintainability.  \\

Reporting the reductions in size as well as change in accuracy for the student model as compared to the corresponding teacher model is useful in our opinion. Although most authors report both these information, some authors don't report either of the two. Moreover, comparing the performance to a trained-from-scratch model of comparable size to the student model is also very informative and we believe should be reported by the authors. \\

Regarding the future of knowledge distillation, most researchers did not provide comments. Nevertheless, Polino et al. \cite{f10} suggested the use of reinforcement learning to enhance development of student models. According to Polino et al., it is not clear how to develop student models that meet memory and processing time constraints. Building a program based on reinforcement learning such that its objective is to optimize memory and processing time requirements would ease development of student models. \\

In addition, most researchers focus on computer vision tasks. For instance, out of the surveyed work, only two considered NLP tasks. Recently, several language models based on transformer architecture \cite{transformer} have been proposed such as Bidirectional Encoder Representations from Transformers (BERT) \cite{BERT}. These models have parameters in the order of hundreds of millions. This issue has motivated several researchers to utilize knowledge distillation \cite{BERT-distill-1, BERT-distill-2}. However, knowledge distillation has not been well investigated yet. Transformer based language models provide better results, in terms of accuracy scores and processing times, than Recurrent Neural Networks (RNNs) \cite{BERT, GPT-2}. As a result, it is important to study knowledge distillation on such models so that relatively small and high performance models could be developed.

\section{Conclusion}
We present several different knowledge distillation methods applied on deep learning architectures. Some of the methods produce more than 80\% decrease in model sizes \cite{f5, f2}. Some other methods provide around 50\% size reductions, but they maintain accuracy scores of teacher models \cite{f10, a5}. In addition, there exist distillation approaches that result in student models with better accuracy scores than their teacher models \cite{a9, f9}. Our criteria are reduction in model sizes and accuracy scores. Consequently, we propose distillation metric which helps in comparing different knowledge distillation algorithms based on their achieved student sizes and accuracy scores. We also highlight different contexts and objectives of some of the knowledge distillation methods, such as limited or absence of the original dataset, improving interpretability, and combining transfer learning with knowledge distillation. \\

Moreover, knowledge distillation is a creative process. There are no rules that guide development of student models or mapping teacher's activations to student's. As a consequence, knowledge distillation highly depends on the domain where it is applied on. Based on requirements of the specific domain, model designers could develop their distillation. We advise designers to focus on simple distillation methods (or build a simpler version of some method) that target a relatively small number of student and teacher layers. This is an important step as it decreases time needed for designers to get familiar with different behaviours of different distillation methods on their domain. After that, they could proceed with more complex methods as they would have developed intuitions about how the methods would behave on their domain of application. As a result, they could eliminate some methods without having to try them. In addition, designers could utilize distillation metric to assess their evaluations.

\section*{Acknowledgment}

The authors would like to thank King Fahd University of Petroleum \& Minerals (KFUPM) for supporting this work.

\bibliographystyle{./bibliography/IEEEtran}
\bibliography{./bibliography/IEEEabrv,./bibliography/References}

\begin{thebibliography}{10}
\providecommand{\url}[1]{#1}
\csname url@samestyle\endcsname
\providecommand{\newblock}{\relax}
\providecommand{\bibinfo}[2]{#2}
\providecommand{\BIBentrySTDinterwordspacing}{\spaceskip=0pt\relax}
\providecommand{\BIBentryALTinterwordstretchfactor}{4}
\providecommand{\BIBentryALTinterwordspacing}{\spaceskip=\fontdimen2\font plus
\BIBentryALTinterwordstretchfactor\fontdimen3\font minus
  \fontdimen4\font\relax}
\providecommand{\BIBforeignlanguage}[2]{{%
\expandafter\ifx\csname l@#1\endcsname\relax
\typeout{** WARNING: IEEEtran.bst: No hyphenation pattern has been}%
\typeout{** loaded for the language `#1'. Using the pattern for}%
\typeout{** the default language instead.}%
\else
\language=\csname l@#1\endcsname
\fi
#2}}
\providecommand{\BIBdecl}{\relax}
\BIBdecl

\bibitem{KD-paper}
G.~Hinton, O.~Vinyals, and J.~Dean, ``Distilling the knowledge in a neural
  network,'' \emph{arXiv preprint arXiv:1503.02531}, 2015.

\bibitem{transfer-learning}
S.~J. Pan and Q.~Yang, ``A survey on transfer learning,'' \emph{IEEE
  Transactions on knowledge and data engineering}, vol.~22, no.~10, pp.
  1345--1359, 2009.

\bibitem{survey}
Y.~Cheng, D.~Wang, P.~Zhou, and T.~Zhang, ``A survey of model compression and
  acceleration for deep neural networks,'' \emph{arXiv preprint
  arXiv:1710.09282}, 2017.

\bibitem{f1}
J.~Yim, D.~Joo, J.~Bae, and J.~Kim, ``A gift from knowledge distillation: Fast
  optimization, network minimization and transfer learning,'' in
  \emph{Proceedings of the IEEE Conference on Computer Vision and Pattern
  Recognition}, 2017, pp. 4133--4141.

\bibitem{a5}
M.~Gao, Y.~Shen, Q.~Li, J.~Yan, L.~Wan, D.~Lin, C.~Change~Loy, and X.~Tang,
  ``An embarrassingly simple approach for knowledge distillation,''
  \emph{arXiv}, pp. arXiv--1812, 2018.

\bibitem{CIFAR-10}
A.~Krizhevsky, G.~Hinton \emph{et~al.}, ``Learning multiple layers of features
  from tiny images,'' 2009.

\bibitem{imagenet}
J.~Deng, W.~Dong, R.~Socher, L.-J. Li, K.~Li, and L.~Fei-Fei, ``Imagenet: A
  large-scale hierarchical image database,'' in \emph{2009 IEEE conference on
  computer vision and pattern recognition}.\hskip 1em plus 0.5em minus
  0.4em\relax Ieee, 2009, pp. 248--255.

\bibitem{CASIA}
D.~Yi, Z.~Lei, S.~Liao, and S.~Z. Li, ``Learning face representation from
  scratch,'' \emph{arXiv preprint arXiv:1411.7923}, 2014.

\bibitem{IJB-A}
B.~F. Klare, B.~Klein, E.~Taborsky, A.~Blanton, J.~Cheney, K.~Allen,
  P.~Grother, A.~Mah, and A.~K. Jain, ``Pushing the frontiers of unconstrained
  face detection and recognition: Iarpa janus benchmark a,'' in
  \emph{Proceedings of the IEEE conference on computer vision and pattern
  recognition}, 2015, pp. 1931--1939.

\bibitem{ResNet}
K.~He, X.~Zhang, S.~Ren, and J.~Sun, ``Deep residual learning for image
  recognition,'' in \emph{Proceedings of the IEEE conference on computer vision
  and pattern recognition}, 2016, pp. 770--778.

\bibitem{a6}
X.~Liu, X.~Wang, and S.~Matwin, ``Improving the interpretability of deep neural
  networks with knowledge distillation,'' in \emph{2018 IEEE International
  Conference on Data Mining Workshops (ICDMW)}.\hskip 1em plus 0.5em minus
  0.4em\relax IEEE, 2018, pp. 905--912.

\bibitem{mnist}
Y.~LeCun, ``The mnist database of handwritten digits,'' \emph{http://yann.
  lecun. com/exdb/mnist/}, 1998.

\bibitem{connect-4}
D.~Dheeru and E.~K. Taniskidou, ``Uci machine learning repository,'' 2017.

\bibitem{f9}
T.~Furlanello, Z.~C. Lipton, M.~Tschannen, L.~Itti, and A.~Anandkumar, ``Born
  again neural networks,'' \emph{arXiv preprint arXiv:1805.04770}, 2018.

\bibitem{DenseNet}
G.~Huang, Z.~Liu, L.~Van Der~Maaten, and K.~Q. Weinberger, ``Densely connected
  convolutional networks,'' in \emph{Proceedings of the IEEE conference on
  computer vision and pattern recognition}, 2017, pp. 4700--4708.

\bibitem{f10}
A.~Polino, R.~Pascanu, and D.~Alistarh, ``Model compression via distillation
  and quantization,'' \emph{arXiv preprint arXiv:1802.05668}, 2018.

\bibitem{quant1}
M.~Courbariaux, Y.~Bengio, and J.-P. David, ``Binaryconnect: Training deep
  neural networks with binary weights during propagations,'' in \emph{Advances
  in neural information processing systems}, 2015, pp. 3123--3131.

\bibitem{Wide-residual-networks}
S.~Zagoruyko and N.~Komodakis, ``Wide residual networks,'' \emph{arXiv preprint
  arXiv:1605.07146}, 2016.

\bibitem{f6}
H.~Wang, D.~Zhang, Y.~Song, S.~Liu, Y.~Wang, D.~Feng, H.~Peng, and W.~Cai,
  ``Segmenting neuronal structure in 3d optical microscope images via knowledge
  distillation with teacher-student network,'' in \emph{2019 IEEE 16th
  International Symposium on Biomedical Imaging (ISBI 2019)}.\hskip 1em plus
  0.5em minus 0.4em\relax IEEE, 2019, pp. 228--231.

\bibitem{U-Net}
{\"O}.~{\c{C}}i{\c{c}}ek, A.~Abdulkadir, S.~S. Lienkamp, T.~Brox, and
  O.~Ronneberger, ``3d u-net: learning dense volumetric segmentation from
  sparse annotation,'' in \emph{International conference on medical image
  computing and computer-assisted intervention}.\hskip 1em plus 0.5em minus
  0.4em\relax Springer, 2016, pp. 424--432.

\bibitem{Janelia}
H.~Peng, M.~Hawrylycz, J.~Roskams, S.~Hill, N.~Spruston, E.~Meijering, and
  G.~A. Ascoli, ``Bigneuron: large-scale 3d neuron reconstruction from optical
  microscopy images,'' \emph{Neuron}, vol.~87, no.~2, pp. 252--256, 2015.

\bibitem{f8}
G.~Kurata and K.~Audhkhasi, ``Improved knowledge distillation from
  bi-directional to uni-directional lstm ctc for end-to-end speech
  recognition,'' in \emph{2018 IEEE Spoken Language Technology Workshop
  (SLT)}.\hskip 1em plus 0.5em minus 0.4em\relax IEEE, 2018, pp. 411--417.

\bibitem{f7}
R.~M. Mun’im, N.~Inoue, and K.~Shinoda, ``Sequence-level knowledge
  distillation for model compression of attention-based sequence-to-sequence
  speech recognition,'' in \emph{ICASSP 2019-2019 IEEE International Conference
  on Acoustics, Speech and Signal Processing (ICASSP)}.\hskip 1em plus 0.5em
  minus 0.4em\relax IEEE, 2019, pp. 6151--6155.

\bibitem{Seq2Seq}
D.~Bahdanau, J.~Chorowski, D.~Serdyuk, P.~Brakel, and Y.~Bengio, ``End-to-end
  attention-based large vocabulary speech recognition,'' in \emph{2016 IEEE
  international conference on acoustics, speech and signal processing
  (ICASSP)}.\hskip 1em plus 0.5em minus 0.4em\relax IEEE, 2016, pp. 4945--4949.

\bibitem{a8}
X.~Jin, B.~Peng, Y.~Wu, Y.~Liu, J.~Liu, D.~Liang, J.~Yan, and X.~Hu,
  ``Knowledge distillation via route constrained optimization,'' in
  \emph{Proceedings of the IEEE International Conference on Computer Vision},
  2019, pp. 1345--1354.

\bibitem{mobilenet}
M.~Sandler, A.~Howard, M.~Zhu, A.~Zhmoginov, and L.-C. Chen, ``Mobilenetv2:
  Inverted residuals and linear bottlenecks,'' in \emph{Proceedings of the IEEE
  conference on computer vision and pattern recognition}, 2018, pp. 4510--4520.

\bibitem{a9}
B.~Heo, M.~Lee, S.~Yun, and J.~Y. Choi, ``Knowledge transfer via distillation
  of activation boundaries formed by hidden neurons,'' in \emph{Proceedings of
  the AAAI Conference on Artificial Intelligence}, vol.~33, 2019, pp.
  3779--3787.

\bibitem{a9.1}
X.~Pan and V.~Srikumar, ``Expressiveness of rectifier networks,'' in
  \emph{International Conference on Machine Learning}, 2016, pp. 2427--2435.

\bibitem{mit_scenes}
A.~Quattoni and A.~Torralba, ``Recognizing indoor scenes,'' in \emph{2009 IEEE
  Conference on Computer Vision and Pattern Recognition}.\hskip 1em plus 0.5em
  minus 0.4em\relax IEEE, 2009, pp. 413--420.

\bibitem{a1}
R.~G. Lopes, S.~Fenu, and T.~Starner, ``Data-free knowledge distillation for
  deep neural networks,'' \emph{arXiv preprint arXiv:1710.07535}, 2017.

\bibitem{celeba}
Z.~Liu, P.~Luo, X.~Wang, and X.~Tang, ``Deep learning face attributes in the
  wild,'' in \emph{Proceedings of the IEEE international conference on computer
  vision}, 2015, pp. 3730--3738.

\bibitem{a2}
A.~Kimura, Z.~Ghahramani, K.~Takeuchi, T.~Iwata, and N.~Ueda, ``Few-shot
  learning of neural networks from scratch by pseudo example optimization,''
  \emph{arXiv preprint arXiv:1802.03039}, 2018.

\bibitem{fashionMNIST}
H.~Xiao, K.~Rasul, and R.~Vollgraf, ``Fashion-mnist: a novel image dataset for
  benchmarking machine learning algorithms,'' \emph{arXiv preprint
  arXiv:1708.07747}, 2017.

\bibitem{a3}
G.~K. Nayak, K.~R. Mopuri, V.~Shaj, R.~V. Babu, and A.~Chakraborty, ``Zero-shot
  knowledge distillation in deep networks,'' \emph{arXiv preprint
  arXiv:1905.08114}, 2019.

\bibitem{a4}
T.~Fukuda, M.~Suzuki, G.~Kurata, S.~Thomas, J.~Cui, and B.~Ramabhadran,
  ``Efficient knowledge distillation from an ensemble of teachers,'' in
  \emph{Interspeech}, 2017, pp. 3697--3701.

\bibitem{a4.2}
Y.~Chebotar and A.~Waters, ``Distilling knowledge from ensembles of neural
  networks for speech recognition.'' in \emph{Interspeech}, 2016, pp.
  3439--3443.

\bibitem{a4.3}
K.~Markov and T.~Matsui, ``Robust speech recognition using generalized
  distillation framework.'' in \emph{Interspeech}, 2016, pp. 2364--2368.

\bibitem{aurora}
H.-G. Hirsch and D.~Pearce, ``The aurora experimental framework for the
  performance evaluation of speech recognition systems under noisy
  conditions,'' in \emph{ASR2000-Automatic Speech Recognition: Challenges for
  the new Millenium ISCA Tutorial and Research Workshop (ITRW)}, 2000.

\bibitem{f5}
S.~L. Pintea, Y.~Liu, and J.~C. van Gemert, ``Recurrent knowledge
  distillation,'' in \emph{2018 25th IEEE International Conference on Image
  Processing (ICIP)}.\hskip 1em plus 0.5em minus 0.4em\relax IEEE, 2018, pp.
  3393--3397.

\bibitem{a7}
G.~Zhou, Y.~Fan, R.~Cui, W.~Bian, X.~Zhu, and K.~Gai, ``Rocket launching: A
  universal and efficient framework for training well-performing light net,''
  in \emph{Thirty-Second AAAI Conference on Artificial Intelligence}, 2018.

\bibitem{svhn}
Y.~Netzer, T.~Wang, A.~Coates, A.~Bissacco, B.~Wu, and A.~Y. Ng, ``Reading
  digits in natural images with unsupervised feature learning,'' 2011.

\bibitem{f2}
T.~He, C.~Shen, Z.~Tian, D.~Gong, C.~Sun, and Y.~Yan, ``Knowledge adaptation
  for efficient semantic segmentation,'' in \emph{Proceedings of the IEEE
  Conference on Computer Vision and Pattern Recognition}, 2019, pp. 578--587.

\bibitem{PASCAL}
M.~Everingham, L.~Van~Gool, C.~K. Williams, J.~Winn, and A.~Zisserman, ``The
  pascal visual object classes (voc) challenge,'' \emph{International journal
  of computer vision}, vol.~88, no.~2, pp. 303--338, 2010.

\bibitem{f4}
M.-C. Wu, C.-T. Chiu, and K.-H. Wu, ``Multi-teacher knowledge distillation for
  compressed video action recognition on deep neural networks,'' in
  \emph{ICASSP 2019-2019 IEEE International Conference on Acoustics, Speech and
  Signal Processing (ICASSP)}.\hskip 1em plus 0.5em minus 0.4em\relax IEEE,
  2019, pp. 2202--2206.

\bibitem{CoViAR}
C.-Y. Wu, M.~Zaheer, H.~Hu, R.~Manmatha, A.~J. Smola, and
  P.~Kr{\"a}henb{\"u}hl, ``Compressed video action recognition,'' in
  \emph{Proceedings of the IEEE Conference on Computer Vision and Pattern
  Recognition}, 2018, pp. 6026--6035.

\bibitem{UCF-101}
K.~Soomro, A.~R. Zamir, and M.~Shah, ``Ucf101: A dataset of 101 human actions
  classes from videos in the wild,'' \emph{arXiv preprint arXiv:1212.0402},
  2012.

\bibitem{HMDB51}
H.~Kuehne, H.~Jhuang, E.~Garrote, T.~Poggio, and T.~Serre, ``Hmdb: a large
  video database for human motion recognition,'' in \emph{2011 International
  Conference on Computer Vision}.\hskip 1em plus 0.5em minus 0.4em\relax IEEE,
  2011, pp. 2556--2563.

\bibitem{a10}
Y.~Liu, J.~Cao, B.~Li, C.~Yuan, W.~Hu, Y.~Li, and Y.~Duan, ``Knowledge
  distillation via instance relationship graph,'' in \emph{Proceedings of the
  IEEE Conference on Computer Vision and Pattern Recognition}, 2019, pp.
  7096--7104.

\bibitem{transformer}
A.~Vaswani, N.~Shazeer, N.~Parmar, J.~Uszkoreit, L.~Jones, A.~N. Gomez,
  {\L}.~Kaiser, and I.~Polosukhin, ``Attention is all you need,'' in
  \emph{Advances in neural information processing systems}, 2017, pp.
  5998--6008.

\bibitem{BERT}
J.~Devlin, M.-W. Chang, K.~Lee, and K.~Toutanova, ``Bert: Pre-training of deep
  bidirectional transformers for language understanding,'' \emph{arXiv preprint
  arXiv:1810.04805}, 2018.

\bibitem{BERT-distill-1}
V.~Sanh, L.~Debut, J.~Chaumond, and T.~Wolf, ``Distilbert, a distilled version
  of bert: smaller, faster, cheaper and lighter,'' \emph{arXiv preprint
  arXiv:1910.01108}, 2019.

\bibitem{BERT-distill-2}
S.~Sun, Y.~Cheng, Z.~Gan, and J.~Liu, ``Patient knowledge distillation for bert
  model compression,'' \emph{arXiv preprint arXiv:1908.09355}, 2019.

\bibitem{GPT-2}
A.~Radford, J.~Wu, R.~Child, D.~Luan, D.~Amodei, and I.~Sutskever, ``Language
  models are unsupervised multitask learners,'' \emph{OpenAI Blog}, vol.~1,
  no.~8, p.~9, 2019.

\end{thebibliography}

\end{document}